\documentclass[11pt,twocolumn]{article}

\usepackage[letterpaper,
            top=1in, bottom=1in,
            left=0.75in, right=0.75in,
            columnsep=0.25in]{geometry}

\usepackage{times}
\usepackage{microtype}
\usepackage{amsmath,amssymb,amsthm}
\usepackage{booktabs}
\usepackage{multirow}
\usepackage{array}
\usepackage{xcolor}
\usepackage{hyperref}
\usepackage{url}
\usepackage{graphicx}
\usepackage{algorithm}
\usepackage{algpseudocode}
\usepackage{enumitem}
\usepackage{caption}
\usepackage{natbib}
\usepackage{fancyhdr}
\usepackage{titlesec}
\usepackage{balance}   

\definecolor{atlasblue}{RGB}{26,58,92}
\definecolor{posblue}{RGB}{26,82,118}
\definecolor{negred}{RGB}{169,50,38}

\hypersetup{colorlinks=true, linkcolor=atlasblue,
            citecolor=atlasblue, urlcolor=atlasblue,
            pdftitle={Compiled Memory: Not More Information, but More Precise Instructions for Language Agents}}

\titleformat{\section}{\normalsize\bfseries}{\thesection}{1em}{}
\titleformat{\subsection}{\normalsize\bfseries\itshape}{\thesubsection}{1em}{}
\titleformat{\subsubsection}{\normalsize\itshape}{\thesubsubsection}{1em}{}
\titlespacing{\section}{0pt}{8pt plus 2pt minus 2pt}{4pt}
\titlespacing{\subsection}{0pt}{6pt plus 1pt minus 1pt}{3pt}

\newcolumntype{L}[1]{>{\raggedright\arraybackslash}p{#1}}
\newcolumntype{C}[1]{>{\centering\arraybackslash}p{#1}}
\newcolumntype{R}[1]{>{\raggedleft\arraybackslash}p{#1}}
\newcommand{\pos}[1]{\textcolor{posblue}{\textbf{#1}}}
\newcommand{\bld}[1]{\textbf{#1}}

\newtheorem{definition}{Definition}

\renewenvironment{abstract}
  {\centerline{\normalsize\bfseries Abstract}\vspace{0.5ex}%
   \begin{quote}\small}
  {\end{quote}}

\begin{document}

\twocolumn[{%
  \centering
  {\LARGE\bfseries Compiled Memory: Not More Information,\\
   but More Precise Instructions for Language Agents\par}
  \vspace{0.75em}
  {\large James Rhodes \quad George Kang\par}
  \vspace{0.2em}
  {\normalsize AlphaBitCore, Inc.\quad
   \texttt{\{james,george\}@alphabitcore.com}\par}
  \vspace{0.2em}
  {\small March 2026\par}
  \vspace{0.75em}
  \begin{abstract}
    Existing memory systems for language agents address \emph{memory
    management}: how to retrieve and page more information within a context
    budget. We address a complementary problem---\emph{memory utility}: what
    experience is worth keeping, and how it should change agent behavior.
    We present \textbf{Atlas}, a memory kernel that compiles accumulated task
    experience into an agent's instruction structure---without fine-tuning,
    RAG, or human intervention. Memory is distillation, not storage; delivery
    is instruction rewriting, not context injection. Facts extracted from agent
    failures and successes are verified through a three-step promotion gate and
    delivered by rewriting the agent's system prompt with learned sub-bullets.

    On CUAD contract analysis, the evolved prompt improves GPT-4o token-level
    F1 by $+8.7$pp and precision by $+12.5$pp. On HotpotQA multi-hop QA,
    joint F1 improves $+3.16$pp. An ablation isolates the mechanism's defining
    property---the \emph{training signal constraint}: the evolved prompt learns
    exactly what it is taught, and nothing more. Applied to Claude Sonnet~4.5
    using the same evolved prompt---compiled from GPT-4o errors,
    unchanged---joint F1 improves $+2.31$pp,
    with gains concentrating where Claude's stronger baseline leaves the most
    room---confirming that the compiled knowledge is task-shaped, not
    model-shaped.
  \end{abstract}
  \vspace{1em}
}]

\section{Introduction}

A fundamental question in agent design is whether an agent can improve its
behavior through accumulated experience. Most existing work addresses
\emph{memory management}: how to retrieve, page, and preserve more
information under a limited context budget---giving the agent access to more
of what it has seen. Storage-first was the right starting point: you cannot
use what you cannot retrieve. But retrieval capacity is now largely a solved
problem, and the harder question has come into focus. Capacity is not the
same as utility. An agent with access to more context does not automatically
know which patterns are trustworthy enough to act on, which failures define
task boundaries, or how to convert that experience into better future
behavior. Recent empirical work confirms that poor memory management
choices---including misaligned experience replay---can actively degrade agent
performance~\cite{xiong2025memory}. More available context can become more
noise.

Atlas addresses the complementary problem: \emph{memory utility}. It asks
what experience is worth promoting into durable memory, and how that
experience should be transformed into instructions that change agent behavior.
The delivery mechanism is instruction rewriting, not context injection. When
an agent repeatedly fails to extract a jurisdiction name because it includes
venue language, the appropriate response is to compile that lesson into the
agent's instruction set as a permanent rule---not inject a reminder as
context. A base agent told to ``extract the EXACT span'' routinely
over-includes context; the same agent with boundary sub-bullets compiled from
training examples learns precisely where to stop. This is closer to program
synthesis than RAG: memory is distillation, not storage.

\paragraph{Contributions.}
\begin{enumerate}[leftmargin=1.2em,itemsep=1pt,topsep=2pt]
  \item \textbf{Atlas architecture}: a four-layer distillation model with
    verified promotion gates and dual-source fact extraction; the read-path
    delivers accumulated experience by rewriting the agent's system prompt
    with learned sub-bullets, not by injecting retrieved context
    (\S\ref{sec:arch}--\ref{sec:evolution}).
  \item \textbf{CUAD}: $+8.7$pp F1 with component ablation (\S\ref{sec:cuad}).
  \item \textbf{HotpotQA}: $+3.16$pp joint F1 with multi-objective training
    (\S\ref{sec:hotpot}).
  \item \textbf{Training signal constraint}: an ablation demonstrating that
    the evolved prompt's ceiling is set precisely by the training signal---adding
    attribution facts widened the attribution gap surgically while leaving the
    answer gap unchanged. The system learned exactly what it was taught, and
    nothing more (\S\ref{sec:ablation}).
  \item \textbf{Cross-model}: Claude Sonnet~4.5 ($n{=}6$) gains $+2.31$pp
    joint F1 from a prompt compiled exclusively from GPT-4o errors, with gains
    concentrating on supporting facts where Claude's stronger answer baseline
    leaves less room to improve. The compiled knowledge is task-shaped, not
    model-shaped (\S\ref{sec:claude}).
\end{enumerate}

Across two structurally different benchmarks and two model families, the same
mechanism produces consistent positive results. The gains are modest ($+3$--$9$pp)
but consistent---evidence for a general mechanism, not a benchmark artifact.

\section{Related Work}

\paragraph{Memory as storage.}
MemGPT~\cite{packer2023memgpt} addresses memory management: hierarchical
tiers plus paging between fast and slow memory allow an agent to operate
beyond its native context window. The framing is virtual context
management---the goal is access to more information. Mem0~\cite{mem0} and
Zep~\cite{zep} add extraction and graph construction but remain
extract-store-retrieve pipelines without verification. \citet{xiong2025memory}
show empirically that addition and deletion choices in such systems affect
downstream behavior in ways that are difficult to predict, with misaligned
experience replay identified as a source of performance degradation. These
systems answer the question: \textit{what should I show the model?} Atlas
answers a different question: \textit{what should the model learn from
experience, and how should that learning alter its instructions?} Where prior
systems improve memory capacity, Atlas addresses memory utility---what to
promote, how to verify it, and how to transform it into behavioral change.

\paragraph{Memory as behavioral signal.}
Reflexion~\cite{shinn2023reflexion} is the closest conceptual ancestor: it
stores verbal feedback in memory with the explicit goal of changing future
agent decisions, not merely expanding context. Atlas extends this intuition
across the full task lifecycle---where Reflexion applies self-evaluation
within a single episode, Atlas promotes verified lessons from many episodes
into durable, structured instructions. ExpeL~\cite{zhao2024expel} takes a
complementary approach: agents gather experiences, extract natural-language
insights, and recall them at inference time to improve performance. The key
distinction from Atlas is delivery: ExpeL injects recalled insights as
context; Atlas compiles verified experience into the agent's instruction
structure, replacing the base prompt permanently at zero additional inference
cost. ExpeL also lacks a verification gate---insights are not checked for
grounding or filtered by confidence before use.

\paragraph{Automatic prompt optimization.}
DSPy~\cite{khattab2024dspy}, OPRO~\cite{yang2024opro}, and
APE~\cite{zhou2023ape} search over prompt variants to maximize a training
metric. Atlas does not search: it annotates a fixed prompt with verified
domain knowledge. The prompt structure is preserved verbatim; only sub-bullets
are added.

\paragraph{Contract understanding.}
CUAD~\cite{hendrycks2021cuad} provides 510 contracts annotated for 41 clause
types. \citet{salminen2025} reports GPT-4o achieves $62.8$ F1 on answerable
CUAD questions---consistent with our $61.9\%$ baseline---while fine-tuned
GPT-4o-mini reaches $82.0$.

\paragraph{Multi-hop QA.}
HotpotQA~\cite{yang2018hotpotqa} requires chaining reasoning across Wikipedia
paragraphs, scoring both answer correctness and supporting fact selection.

\section{Architecture}
\label{sec:arch}

\subsection{Four-Layer Distillation Model}

Atlas organizes memory into four layers of increasing epistemic maturity.
The innermost two---\textbf{Fresh} (ephemeral, run-scoped) and \textbf{Task}
(durable, workspace-scoped)---accumulate raw signal within a run or task.
The outer two are durable and long-lived: \textbf{Contextual}
(workspace-scoped) stores one episode per completed task;
\textbf{Historical} (tenant-scoped) holds verified facts with
\texttt{factKey}, confidence scores, validity windows, and corroboration
counts. Only Historical-layer facts reach the evolved prompt.

Promotion is event-driven: \texttt{end\_run} executes Fresh$\!\to$Task,
Task$\!\to$Contextual (episode creation), and Contextual$\!\to$Historical
(three-step gate).

\subsection{Three-Step Verification Gate}

\textbf{Step~1 (deduplication):} candidates embedded and queried; matches
above $\tau_d{=}0.92$ rejected. \textbf{Step~2 (LLM verification):}
non-duplicate candidates evaluated in batches of $B{=}10$; verdicts:
\texttt{accept}, \texttt{reject}, \texttt{merge}, \texttt{needs\_review}.
On LLM failure, $v\!\leftarrow\!\texttt{needs\_review}$---never silent
promotion. \textbf{Step~3 (grounding):} the candidate must share at least
two significant keywords with the source episode---a lexical heuristic that
filters fabricated facts without requiring semantic entailment
(\S\ref{sec:limits}).

\subsection{Convergence and Confidence}

\begin{definition}[Confidence Process]
Let $c_0{=}0.5$. After event $i$:
$c_{i+1} = 1-(1-c_i)\lambda_+$ (corroboration) or
$c_{i+1} = c_i\cdot\lambda_-$ (contradiction),
with $\lambda_+{=}0.3$, $\lambda_-{=}0.7$.
\end{definition}

From $c_0{=}0.5$: one corroboration $\to c{=}0.85$; two $\to c{=}0.955$.
A contradiction from $c{=}0.95$ yields $c{=}0.665$, flagging for review.

\subsection{Learning from Failures and Successes}
\label{sec:failures_successes}

Corroboration enriches fact content. When a corroborating candidate contains
boundary rules or trigger phrases absent from the existing fact, an LLM
merges the new specifics in. Fact richness $\rho(f)$ is monotonically
non-decreasing.

The training pipeline learns from \emph{both} directions. Agent errors produce
\textbf{boundary rules}---where to start and stop. Agent successes on hard
cases produce \textbf{guard facts}---descriptions of correct behavior that
protect rules from being weakened during evolution.

\paragraph{Example (Governing Law, CUAD).}
The agent extracts ``...the laws of the State of New York, and any disputes
shall be resolved in the courts of Manhattan'' when gold is only ``the laws of
the State of New York.'' The analysis LLM produces a boundary rule:
\textit{extract only the jurisdiction name; stop before venue clauses.} After
10 corroborations, later correct extractions generate a guard fact: \textit{The
agent correctly extracted the jurisdiction name without including venue language.}
Both reach the evolved prompt: the rule instructs where to stop; the guard
prevents the evolution LLM from weakening it.

\section{Prompt Evolution}
\label{sec:evolution}

Context injection treats memory as data to reason over. Prompt evolution
treats memory as compiled instructions. The evolved prompt \emph{replaces}
the base prompt---same inference cost, no additional context at runtime.

\subsection{Pipeline}

\textbf{Retrieve and filter.} Top facts filtered to $c\geq 0.85$,
deduplicated to one fact per clause/question type, sanitized to remove
instance-specific entity names.

\textbf{Evolve.} An LLM adds facts as sub-bullets under relevant instruction
lines. Mandatory coverage ensures all selected facts appear. Original prompt
text preserved verbatim.

\textbf{Anchor.} Strategically positioned instructions counterbalance the
dominant failure direction: recall anchors for precision-oriented tasks
(CUAD), accuracy anchors for reasoning tasks (HotpotQA).

\subsection{Training Signal Constraint}
\label{sec:constraint}

The evolved prompt can only be as good as the facts it draws from. If the
training signal captures only one dimension of agent behavior, the evolved
prompt improves only that dimension. This is an architectural property, not a
bug: the read-path output is bounded by the write-path input. Practitioners
deploying Atlas should characterize all evaluation dimensions of their task
and ensure the analysis prompt extracts facts covering each. The constraint is
directly demonstrated by the HotpotQA ablation (\S\ref{sec:ablation}).

\begin{algorithm}[t]
\caption{Write Path}
\label{alg:write}
\small
\begin{algorithmic}[1]
\Require Examples $(x_i,y_i)$; fact store $\mathcal{F}$
\For{all $(x_i, y_i)$}
  \State Run agent on $x_i$; compare to gold $y_i$
  \State Extract: boundary rules (errors), guard facts (successes)
  \For{each fact $c$} \Comment{3-step gate}
    \If{dup \textbf{or} unverified \textbf{or} ungrounded}
      quarantine
    \ElsIf{$\exists f\in\mathcal{F}: f.\text{key}=c.\text{key}$}
      classify \& enhance if corroboration
    \Else~promote $c$ to $\mathcal{F}$
    \EndIf
  \EndFor
\EndFor
\end{algorithmic}
\end{algorithm}

\begin{algorithm}[t]
\caption{Read Path (Prompt Evolution)}
\label{alg:read}
\small
\begin{algorithmic}[1]
\Require Base prompt $P_0$; fact store $\mathcal{F}$
\State Filter: $c\geq0.85$, dedup by type, sanitize
\State $P_1\leftarrow\text{LLM}_\text{evolve}(P_0,\text{facts})$
\State $P_2\leftarrow\text{inject\_anchors}(P_1)$
\State\Return $P_2$ \Comment{Replaces $P_0$ at inference}
\end{algorithmic}
\end{algorithm}

\section{Evaluation I: CUAD}
\label{sec:cuad}

\subsection{Setup}

\textbf{Dataset:} CUAD~\cite{hendrycks2021cuad}, 510 contracts, 41 clause
types (SQuAD 2.0 format). Positive examples only (base handles negatives
well; 86\% negative class would inflate all metrics). \textbf{Knowledge
base:} 1,140 training docs $\to$ 338 facts $\to$ 25 selected ($c\geq0.85$,
23 clause types). \textbf{Design:} \{base, evolved\}$\times$6 runs, GPT-4o,
temperature 0, $n{=}500$ per run.

\subsection{Results}

Table~\ref{tab:cuad} reports per-example paired $t$-tests pooled across
6 runs ($n{=}3{,}000$ observations). The evolved prompt wins all 6/6 runs
on token F1 (range: $+7.5$pp to $+10.7$pp).

\begin{table}[t]
\centering
\caption{GPT-4o on CUAD ($n{=}3{,}000$ paired observations).}
\label{tab:cuad}
\small\setlength{\tabcolsep}{4pt}
\begin{tabular}{lrrr}
\toprule
\textbf{Metric} & \textbf{Evo} & \textbf{Base} & $\boldsymbol{\Delta}$ \\
\midrule
\bld{Token F1}     & $70.5\%$ & $61.9\%$ & \pos{$+8.7$pp}  \\
Exact Match        & $36.9\%$ & $32.4\%$ & $+4.5$pp  \\
Precision          & $71.0\%$ & $58.5\%$ & \pos{$+12.5$pp} \\
Recall             & $81.3\%$ & $82.6\%$ & $-1.3$pp  \\
FN Rate            & $6.6\%$  & $9.6\%$  & $-3.0$pp  \\
\bottomrule
\multicolumn{4}{l}{\scriptsize All $|\Delta|>1$pp: $p<10^{-7}$. F1: $d=0.27$.}
\end{tabular}
\end{table}

The gain is driven by precision ($+12.5$pp): boundary rules teach the agent
where to stop, reducing over-inclusion. Recall is held within $1.3$pp by
recall anchors. The false negative rate improves ($-3.0$pp)---the evolved
agent finds more clauses while extracting more precise spans.\footnote{The
$61.9\%$ baseline is consistent with independently reported GPT-4o performance
($62.8\%$; \citealt{salminen2025}).}

\subsection{Component Ablation}

\begin{table}[t]
\centering
\caption{CUAD component ablation ($n{=}1$, 200 examples/variant).}
\label{tab:cuad_ablation}
\small\setlength{\tabcolsep}{3pt}
\begin{tabular}{L{2.8cm}rrrr}
\toprule
\textbf{Config} & \textbf{Docs} & \textbf{Facts} & $\boldsymbol{\Delta}$\textbf{F1} \\
\midrule
50 docs, no dedup      & 50    & 4     & $+6.7$pp \\
200 docs, no dedup     & 200   & 13    & $+5.5$pp \\
400 + C1 only          & 400   & 20    & $+1.0$pp \\
C1+C2 (quality only)   & 1,140 & 7     & $+4.4$pp \\
\bld{C1--C5 (full)}    & 1,140 & 23    & \pos{$+9.1$pp} \\
\bottomrule
\end{tabular}
\end{table}

Two findings from this ablation recur in the HotpotQA results. \textbf{(1) More data without quality
controls is worse than less data with clean facts}: 4 clean facts ($+6.7$pp)
outperform 20 noisy facts ($+1.0$pp). \textbf{(2) Precision rules without
recall countermeasures collapse recall} ($-9.6$pp at C1+C2): mandatory
coverage (C3) + recall anchors (C5) recovered recall while further improving
precision.

\section{Evaluation II: HotpotQA}
\label{sec:hotpot}

\subsection{Dataset and Setup}

HotpotQA~\cite{yang2018hotpotqa}: multi-hop QA over Wikipedia paragraphs.
Each example: question, 10 context paragraphs (2 gold + 8 distractor), gold
answer span, gold supporting facts ($\sim$75\% bridge, $\sim$25\% comparison).
\textbf{Primary metric:} Joint F1 (answer\_f1 $\times$ supp\_fact\_f1)---penalizes
metric displacement. \textbf{Design:} \{base, evolved\}$\times$6 runs, GPT-4o,
500 stratified examples/run (375 bridge + 125 comparison).

\subsection{Multi-Objective Fact Extraction}

The training signal constraint is directly observable here. An initial
implementation (\textbf{v2}) extracted facts about answer derivation only.
Answer F1 improved; supporting-fact F1 degraded---the restructured reasoning
had decoupled answer derivation from sentence-level attribution, and the
training signal had not been told to care.

The \textbf{v3} analysis prompt requires facts about both dimensions
(Table~\ref{tab:multiobjective}) and states explicitly: \textit{``Improving
the answer while degrading supporting fact selection is NOT a net win.''} This
produced 45 facts, 18 selected for evolution.

\begin{table}[t]
\centering
\caption{Multi-objective fact extraction categories (v3).}
\label{tab:multiobjective}
\small\setlength{\tabcolsep}{4pt}
\begin{tabular}{L{1.3cm}L{4.2cm}}
\toprule
\textbf{Dim.} & \textbf{Example Rule} \\
\midrule
ANSWER      & ``Extract BOTH birth dates; do not infer from paragraph order'' \\[3pt]
ATTRIB.     & ``Include the sentence with the compared value, not intro sentences'' \\[3pt]
BOTH        & ``Extract from paragraph B AND cite sentences from both A and B'' \\
\bottomrule
\end{tabular}
\end{table}

\subsection{Results}

Table~\ref{tab:hotpot} reports per-example paired $t$-tests pooled across
6 runs ($n{=}3{,}000$ observations). Every metric improves; 6/6 run wins on
joint F1 (range: $+2.1$pp to $+4.6$pp).

\begin{table}[t]
\centering
\caption{GPT-4o v3 on HotpotQA ($n{=}3{,}000$ paired observations).}
\label{tab:hotpot}
\small\setlength{\tabcolsep}{4pt}
\begin{tabular}{lrrr}
\toprule
\textbf{Metric} & \textbf{Evo} & \textbf{Base} & $\boldsymbol{\Delta}$ \\
\midrule
\bld{Joint F1}  & $51.80\%$ & $48.64\%$ & \pos{$+3.16$pp} \\
Joint EM        & $23.33\%$ & $19.67\%$ & $+3.67$pp \\
Answer F1       & $79.27\%$ & $78.07\%$ & $+1.21$pp \\
Answer EM       & $65.90\%$ & $64.67\%$ & $+1.23$pp \\
Supp.\ Fact F1  & $63.18\%$ & $60.01\%$ & \pos{$+3.17$pp} \\
Supp.\ Fact EM  & $32.50\%$ & $28.03\%$ & $+4.47$pp \\
\bottomrule
\multicolumn{4}{l}{\scriptsize All metrics: $p<10^{-5}$. Joint F1: $d=0.145$.}
\end{tabular}
\end{table}

\paragraph{Note on effect sizes.} $d\approx0.14$ is modest by Cohen's
conventions; the extreme $p$-values reflect 3,000 paired observations, not an
unusually large effect. The claim is $+3$pp joint F1 with no human
intervention---evidence for the mechanism, not a state-of-the-art claim.

\section{The Training Signal Constraint}
\label{sec:ablation}

The training signal constraint can be observed directly. Table~\ref{tab:ablation}
compares v2 (answer-only facts) against v3 (multi-objective facts) on
identical test sets and base models---only the analysis prompt changed.
The base prompts are statistically identical across versions (answer F1:
$78.06\%$ vs.\ $78.07\%$; supp.\ fact F1: $60.54\%$ vs.\ $60.01\%$).

\begin{table}[t]
\centering
\caption{v2 vs.\ v3 ablation: only the training signal changed.}
\label{tab:ablation}
\small\setlength{\tabcolsep}{4pt}
\begin{tabular}{lrrr}
\toprule
\textbf{Metric} & \textbf{v2 $\Delta$} & \textbf{v3 $\Delta$} & \textbf{Shift} \\
\midrule
Answer F1        & $+1.28$pp & $+1.21$pp & $-0.07$ (flat)   \\
Answer EM        & $+1.00$pp & $+1.23$pp & $+0.23$ (flat)   \\
Supp.\ Fact F1   & $+2.35$pp & \pos{$+3.17$pp} & \pos{$+0.82$pp} \\
Supp.\ Fact EM   & $+3.13$pp & \pos{$+4.47$pp} & \pos{$+1.34$pp} \\
\bld{Joint F1}   & $+2.13$pp & \pos{$+3.16$pp} & \pos{$+1.03$pp} \\
\bottomrule
\end{tabular}
\end{table}

The result is surgical: adding attribution facts specifically widened the
supporting-fact gap while leaving the answer gap essentially unchanged.
The system learned precisely what it was taught---and nothing more.

This is the central property of compiled memory: the read-path output is
bounded by the write-path input. The training signal constraint is not a
limitation to work around---it \emph{is} the mechanism.

\section{Cross-Model Generalization}
\label{sec:claude}

Table~\ref{tab:claude} reports Claude Sonnet~4.5 on HotpotQA v3
($n{=}6$ runs, $3{,}000$ paired observations). The same evolved
prompt compiled from GPT-4o errors is used without modification.

\begin{table}[t]
\centering
\caption{Claude Sonnet~4.5 vs.\ GPT-4o v3 on HotpotQA.
$n{=}6$ runs $\times$ 500 examples $= 3{,}000$ paired observations
(Claude); $n{=}6$ (GPT-4o). Same evolved prompt for both---compiled
exclusively from GPT-4o training errors.}
\label{tab:claude}
\small\setlength{\tabcolsep}{4pt}
\begin{tabular}{lrrrr}
\toprule
\textbf{Metric} & \textbf{Cl.\ Evo} & \textbf{Cl.\ Base} & \textbf{Cl.\ $\Delta$} & \textbf{GPT $\Delta$} \\
\midrule
\bld{Joint F1}  & $54.93\%$ & $52.61\%$ & \pos{$+2.31$pp} & $+3.16$pp \\
Joint EM        & $28.90\%$ & $26.43\%$ & $+2.47$pp & $+3.67$pp \\
Answer F1       & $82.01\%$ & $81.69\%$ & $+0.32$pp & $+1.21$pp \\
Supp.\ Fact F1  & $65.47\%$ & $63.10\%$ & \pos{$+2.37$pp} & $+3.17$pp \\
Supp.\ Fact EM  & $40.10\%$ & $36.73\%$ & $+3.37$pp & $+4.47$pp \\
\bottomrule
\multicolumn{5}{l}{\scriptsize Joint F1: $p{<}0.001$, $d{=}0.067$. Answer F1: $p{=}0.22$ (ns).}\\
\multicolumn{5}{l}{\scriptsize Supp.\ Fact F1: $p{<}0.001$. 6/6 run wins on joint F1.}
\end{tabular}
\end{table}

\paragraph{The capability-gap effect in cross-model data.}
Claude's baseline answer F1 ($81.69\%$) is already $3.6$pp above
GPT-4o's ($78.07\%$). Consistent with the capability-gap prediction,
the evolved prompt adds nothing to Claude's answer quality
($+0.32$pp, $p{=}0.22$, ns)---there is almost no gap to fill. But
supporting fact quality, where Claude still has room to improve, gains
$+2.37$pp ($p{<}0.001$). Joint F1 improves $+2.31$pp on all 6/6 runs.

Uniform gains across both models would suggest model-specific prompt tuning.
Concentrated gains---where each model improves most where it had the most
room---suggest something different. The same evolved prompt---compiled from
GPT-4o errors only---benefits each model differently, concentrating where
each has room to improve.
This confirms that the compiled knowledge is \emph{task-shaped, not
model-shaped}: the instruction set encodes what the task requires, and
each model extracts benefit proportional to its own remaining gap.

\section{Cross-Domain Analysis}

CUAD and HotpotQA differ on nearly every axis: task type (span extraction
vs.\ multi-hop QA), error mode (over-inclusion vs.\ missed hops), compiled
fact type (boundary rules vs.\ reasoning chain rules), and primary gain
($+8.7$pp precision-driven vs.\ $+3.16$pp balanced joint F1). Despite these
differences, the same mechanism produces consistent positive results. That
consistency is the strongest evidence for generality: the claim is not that
Atlas is optimal for any individual task, but that verified experience
compilation is a general mechanism, bounded in each case by the training
signal constraint (\S\ref{sec:constraint}).

\section{Limitations}
\label{sec:limits}

\paragraph{Positive-only (CUAD).} Recall anchors may increase false
positives on negative examples; full evaluation needed.

\paragraph{Ablation design.} The CUAD component ablation (Table~\ref{tab:cuad_ablation}) runs $n{=}1$/variant; formal ablations with $n\geq3$ are needed to tighten those estimates. The v2 vs.\ v3 training signal ablation (Table~\ref{tab:ablation}) uses the full $n{=}6$ run design ($3{,}000$ paired observations per version) and does not share this limitation.

\paragraph{Grounding heuristic.} $k_{\min}{=}2$ keyword overlap is not
semantic entailment; ``verified'' describes the full three-step gate.

\paragraph{Token budget.} The evolved prompt is $\sim$5$\times$ longer than
the base prompt ($\sim$960$\to$$\sim$1,570 tokens). Whether gains reflect
content specificity or simply additional tokens is unresolved; a
length-matched control with generic content would isolate the two effects.

\paragraph{Analysis LLM calibration.} Extraction errors become compiled
instructions; human review of sampled facts is planned.

\section{Conclusion}

We have asked whether agents can teach themselves. The evidence across two
structurally different benchmarks and two model families suggests yes, in a
meaningful and bounded sense. On CUAD, compiled boundary rules improve GPT-4o
F1 by $+8.7$pp. On HotpotQA, compiled reasoning and attribution rules improve
joint F1 by $+3.16$pp. Claude Sonnet~4.5 ($n{=}6$, same evolved prompt) shows $+2.31$pp
joint F1 from a prompt compiled entirely from GPT-4o errors.

The mechanism's defining property is the training signal constraint. Targeting
answer derivation alone improved answers but degraded attribution; targeting
both recovered the regression precisely. The system learned exactly what it
was taught---and nothing more. This is not a limitation; it is the mechanism.

The gains are modest. They are not competitive with fine-tuning. But they
require no labeled data, no gradient updates, and no prompt engineering.
An agent accumulates verified experience; that experience compiles
automatically into more precise instructions; the next run performs better.
That is what it means, in a practical sense, for an agent to teach itself.

\balance
\bibliographystyle{plainnat}

\end{document}